# SNAPO: Smooth Neural Adjoint Policy Optimization for Optimal Control via Differentiable Simulation

**Dmitri Goloubentsev** and **Natalija Karpichina** MatLogica, U.K.

*Many real-world problems require sequential decisions under uncertainty: when to inject or withdraw gas from storage, how to rebalance a pension portfolio each month, what temperature profile to run through a pharmaceutical reactor chain. Dynamic programming solves small instances exactly but scales exponentially in state dimensions. Black-box reinforcement learning handles high-dimensional states but trains slowly and produces no sensitivities. We introduce SNAPO (Smooth Neural Adjoint Policy Optimization), a framework that embeds a neural policy inside a known, differentiable simulator, replaces hard constraints with smooth approximations, and computes exact gradients of the objective with respect to all policy parameters and all inputs in a single adjoint pass. We demonstrate SNAPO on three domains: natural gas storage (training in under a minute, 365 forward curve sensitivities at no additional cost per sensitivity), pension fund asset-liability management (6.5x–200x sensitivity speedup over bump-and-revalue, scaling with the number of risk factors), and pharmaceutical manufacturing (cross-unit sensitivities through a 4-unit process chain, with 20 ICH Q8 regulatory sensitivities from 5 adjoint passes in 74.5 milliseconds). All sensitivities are produced by the same backward pass that trains the policy, at a cost proportional to one reverse pass regardless of how many sensitivities are computed.*

## 1. Introduction

Consider a gas storage operator who decides daily how much gas to inject or withdraw, watching inventory and spot price. Or a pension fund manager rebalancing a three-asset portfolio every month against shifting rates and funded ratios. Or a pharmaceutical engineer adjusting cooling rates through a reactor chain to hit regulatory quality targets. Different industries, same mathematical structure: maximize a cumulative objective through a sequence of constrained decisions under uncertainty.

Dynamic programming (DP) computes exact optima by discretizing the state space and solving backward. For a gas storage facility with 150 inventory levels and 20 demand bins, this means 3,000 states per day — tractable for one facility, exponential for a portfolio.[1] Sensitivity computation faces a different scaling problem: a pension fund with 200 risk factors needs 201 forward evaluations per sensitivity report via bump-and-revalue, and a pharmaceutical chain with 7 process parameters and 10 quality attributes needs 14 finite-difference evaluations to fill the ICH Q8 sensitivity matrix.

Black-box reinforcement learning (PPO, SAC) handles high-dimensional state spaces but treats the simulator as opaque. Gradients are estimated from scalar rewards over long episodes, producing high-variance updates. Training a gas storage policy takes 3.7 hours via SAC.[2] No sensitivities are produced.

When the dynamics are known — as they are in all three cases — we can differentiate directly through the simulator. SNAPO embeds a small neural network inside the simulation as connective tissue between known physical processes. The entire forward computation (state dynamics, policy decisions, constraint enforcement, reward calculation) is recorded as a single differentiable program. One adjoint (reverse-mode) pass then yields the gradient of the objective with respect to every policy weight and every input simultaneously. The cost of this pass is approximately equal to one forward evaluation, regardless of the number of parameters.[3]

We demonstrate this framework on three domains spanning commodity trading, pension management, and pharmaceutical manufacturing. In each case, SNAPO trains in seconds to minutes, matches or exceeds domain-specific baselines, and produces all required sensitivities at constant cost from the same adjoint pass. The domains share a common structure (state dynamics, neural policy, smooth constraints, adjoint training) and differ only in the physics and the objective.

## 2. Related work

**Adjoint methods for sensitivity computation.** Reverse-mode automatic differentiation has long been used for computing Greeks in derivatives pricing (Giles and Glasserman, 2006; Capriotti et al., 2017) and for training continuous-depth neural networks via the adjoint sensitivity method (Chen et al., 2018). Huge and Savine (2020) use adjoint-generated labels to train neural approximators of pricing functions. These approaches compute sensitivities of a fixed computation; SNAPO extends the adjoint to optimize the computation itself — training a policy embedded within the simulator.

**Differentiable simulation for control.** de Avila Belbute-Peres et al. (2018) demonstrated end-to-end learning through differentiable physics engines. Hu et al. (2020) introduced DiffTaichi, a differentiable programming system for physical simulation that records computations on tape and replays them in reverse, achieving 188x faster convergence than reinforcement learning on robotics tasks. Brax (Freeman et al., 2021) provides a differentiable rigid-body physics engine with an analytic policy gradient mode. Mora et al. (2021) train neural policies via differentiable simulation in PODS, the closest architectural precedent to SNAPO. Both embed policies inside differentiable simulators and train via analytic gradients. Where PODS targets deterministic rigid-body physics, SNAPO targets stochastic financial and industrial simulation, where contract constraints are non-differentiable and require explicit smooth relaxation. SNAPO also uses the adjoint pass to produce input sensitivities (Greeks, risk factors, regulatory metrics) alongside policy gradients, which PODS does not. Xu et al. (2022) propose Short-Horizon Actor-Critic (SHAC), which backpropagates exact gradients through short windows of differentiable simulation and uses a learned critic for long-horizon credit assignment. Suh et al. (2022) analyze when first-order gradients from differentiable simulators are reliable, showing that discontinuities and stiffness can compromise gradient quality — the problem SNAPO's smooth approximations address.

**Model-based reinforcement learning.** DreamerV3 (Hafner et al., 2025) learns a latent world model and propagates analytic gradients through imagined trajectories. When the simulator is known, as in all three domains here, learning the dynamics wastes data and introduces approximation error. Donti et al. (2017) differentiate through convex optimization for energy storage arbitrage — a decision-focused approach that SNAPO generalizes to non-convex, stochastic simulation.

**Domain-specific prior art.** For gas storage: Henaff (2012) applied adjoint AD to gas storage hedging — computing storage sensitivities, but not optimizing a neural policy within the simulation. Boogert and de Jong (2008) apply LSMC; Thompson et al. (2009) solve by backward DP; Carmona and Ludkovski (2010) formulate storage as optimal switching; Chen and Forsyth (2007) use semi-Lagrangian methods; Curin et al. (2021) train deep learning strategy networks from LSMC outputs; Balaconi et al. (2025) train SAC agents on a market simulator. Warin (2012) developed a multi-factor adjoint approach for storage hedging, and Denkert, Pham, and Warin (2024) extend the optimal-switching formulation to policy gradient via control randomisation, with energy real-options case studies — the closest published work on the policy-gradient side of SNAPO. For insurance ALM: Schlögl (2022) applies operator-overloading AAD to actuarial models; Cathcart et al. (2023) use forward-mode AD for actuarial sensitivity analysis; Krah et al. (2020) train proxy models achieving 2-4% VaR error. For pharmaceutical process optimization: PharmaPy (Lakerveld et al.) and gPROMS provide simulation environments but do not optimize embedded neural controllers via adjoint differentiation.

The architecture of embedding a neural policy inside a differentiable simulator and training via analytic gradients has been demonstrated for robotics (PODS, Brax, DiffTaichi). SNAPO's two specific contributions relative to this prior work are: (1) a systematic smooth relaxation of non-differentiable contract constraints (min/max inventory, regulatory limits, if-else logic) that enables gradient flow through constraints that would otherwise block it, with explicit quantification of the resulting bias; and (2) dual use of the adjoint pass, where the same reverse computation that trains the policy also produces all input sensitivities (forward curve Greeks, risk factor sensitivities, regulatory metrics) as a primary deliverable rather than a byproduct.

# 3. Method

## *3.1 Problem formulation*

We consider sequential decision problems of the form:

$$\max_{\theta} \ \mathbb{E}\left[\sum_{t=0}^{T} r(x_t, u_t, w_t)\right]$$

where $x_t$ is the state (inventory, funded ratio, concentrations), $u_t$ is the control action (injection rate, portfolio weights, temperature setpoint), $w_t$ is exogenous uncertainty (price shocks, market returns, measurement noise), and the dynamics $x_{t+1} = f(x_t, u_t, w_t)$ are known. Box constraints $g(x_t, u_t) \leq 0$ enforce physical and regulatory limits.

### *3.2 Neural policy*

The control action is parameterized as $u_t = \pi(x_t; \theta)$, where $\pi$ is a small multilayer perceptron (MLP), typically two hidden layers with 5-24 units and tanh activations. For problems with strong temporal patterns (gas storage seasonality, monthly ALM rebalancing), we add learnable time-dependent biases $b_t$ that capture periodicity without requiring the MLP to learn it from inputs.

The 68-822 parameter range suits the compiled-adjoint regime: a JIT-compiled AVX2 reverse pass at this size runs in single-digit to low-hundreds of milliseconds, so per-iteration cost is dominated by the forward simulation rather than by gradient assembly. This favours simulator-driven structure over policy-network capacity and is the regime where a compiled adjoint is most economical relative to interpreted alternatives.

The policy sits inside the simulator as a component of the forward computation, not as a separate controller that queries the simulator. In black-box RL, the gradient stops at the reward signal. Here, it flows through the entire chain: state dynamics, policy evaluation, constraint enforcement, and reward calculation.

### *3.3 Smooth constraint relaxation*

Hard constraints (min, max, if-else) are non-differentiable. We replace them with smooth approximations:

$$\text{smooth_max}(a, b) = \frac{1}{2}\left(a + b + \sqrt{(a-b)^2 + \frac{1}{k}}\right)$$

with sharpness parameter $k = 50$. Penalty terms enforce soft boundaries: $\text{penalty} = \lambda \cdot \text{smooth_relu}(\text{violation})^2$.

This introduces bias: the smooth objective differs from the hard-constraint objective, and we quantify the cost per domain in Section 6.1. The tradeoff is explicit. Without smoothing, no gradients cross the constraint boundary. With smoothing, they do, but the optima shift.

### *3.4 Tape recording and adjoint pass*

The entire forward simulation (dynamics, policy, constraints, rewards) for $N$ Monte Carlo paths over $T$ time steps is recorded as a single computation graph.

This recording happens once; subsequent evaluations replay the graph with different random seeds and input values.

One forward replay computes $J = \frac{1}{N}\sum_{n=1}^{N}\sum_{t=0}^{T} r_t^{(n)}$. One adjoint (reverse) replay then yields:

- $\frac{\partial J}{\partial \theta}$: gradients with respect to all policy parameters (for training)
- $\frac{\partial J}{\partial \text{inputs}}$: sensitivities to all model inputs (at no additional cost per input)

The cost of the reverse pass is approximately 1-2x the forward pass, independent of the number of parameters or inputs.[4] $N$ sensitivities cost one reverse pass, not $N$ forward evaluations.

### *3.5 Training*

We optimize $\theta$ via Adam (Kingma and Ba, 2015) on the adjoint gradients:

```
Record simulation + policy + constraints as graph G
for iter = 1 to K:
    Forward: replay G on N Monte Carlo paths -> J
    Reverse: adjoint pass through G -> dJ/d(theta), dJ/d(inputs)
    theta <- Adam(theta, dJ/d(theta), lr=0.005)
Return trained policy, sensitivity function
```

Typical settings: $N = 256\text{-}2{,}048$ paths, $K = 500\text{-}1{,}000$ iterations, learning rate 0.005. Total training time: 5 seconds to 95 seconds across all domains, on a single CPU core with no GPU.

Adam is sufficient for the parameter regime studied here and is used throughout for cross-domain consistency. Second-order methods (TRON, Levenberg-Marquardt) and global-search methods (Differential Evolution, Evolution Strategies) are well-suited to small-parameter problems but introduce trade-offs in iteration count and forward-evaluation budget; we discuss alternatives in Section 9.

## 4. Experimental setup

All experiments run on Linux 6.1.0-13-amd64 (Intel Xeon Platinum 8280L), compiled with AVX2, single-threaded unless noted, no GPU. The adjoint computation uses a compiled kernel produced by a commercial AD library.[5]

### *4.1 Gas storage*

A gas storage facility makes daily inject/withdraw decisions over a 365-day horizon under stochastic prices. We use the Schwartz-Smith two-factor model calibrated to six Henry Hub forward curves. The policy is a 2-layer MLP (5+5 hidden, tanh, 2 outputs) with 730 daily biases, totaling 822 parameters (92 MLP weights + 730 biases). Training uses 256 Monte Carlo paths; out-of-sample evaluation uses 1,024 separate paths (seed 999). Baselines: backward DP (deterministic and stochastic demand), LSMC, PPO, SAC. Full details in the companion paper (Karpichina, 2026).

### *4.2 Insurance ALM*

A pension fund with three asset classes (equity, government bond, credit) rebalances monthly over a 10-year projection. Equity follows GBM, the short rate follows Hull-White, credit spreads follow OU. The policy is a 3-layer MLP (5-5-5-2) with 240 monthly biases, totaling 312 parameters. We test on two real pension fund calibrations: a CalPERS-like fund (\$556B AUM, 79% funded ratio, 6.8% discount rate) and an Ontario Teachers-like fund (\$279B AUM, 111% funded, 5.5% discount). Training uses 2,048 scenarios. Baselines: bump-and-revalue (N+1 forward evaluations for N sensitivities), static allocation strategies (60/40, 70/30, 30/70).

### *4.3 Pharmaceutical manufacturing*

A 4-unit process chain (continuous stirred-tank reactor, batch crystallizer, cake filter, falling-rate dryer) converts raw materials to dried pharmaceutical product. The neural controller sets cooling rate and gas temperature via a 2-layer MLP (8-6-2, 68 parameters). The simulation integrates coupled ODEs (Arrhenius kinetics, method-of-moments crystallization, Darcy filtration, falling-rate drying) over 9,300 Euler steps. ICH Q8 sensitivities — how each Critical Quality Attribute (CQA) responds to each Critical Process Parameter (CPP) — are computed from adjoint passes. Baselines: finite-difference sensitivities, PharmaPy (Lakerveld et al.) as Python reference implementation.

## 5. Results

### *5.1 Gas storage*

Table 1: SNAPO vs baselines on seasonal Henry Hub curve (Schwartz-Smith)

| Method | Value | Training | Eval (1,024 paths) | Greeks |
|---|---|---|---|---|
| DP (stochastic demand) | 2.47M | 1.1 s | < 0.1 ms | 5.1 s (bump) |
| SNAPO (9-input) | 2.35M | 47 s | 0.1 ms | One adjoint pass |
| SAC | 1.07M | ~14,000 s | ~5 ms | None |
| PPO | 1.41M | ~206 s | ~5 ms | None |

Table 1 shows the seasonal curve, where DP wins by 5% (2.47M vs 2.35M). LSMC with polynomial basis collapsed under stochastic demand and is omitted; neural regression (Becker et al., 2019) or parametric policy optimization (Curin et al., 2021) could perform better but were not tested.[6]

SNAPO and DP optimize different objectives: SNAPO maximizes the smoothed problem (differentiable constraints, k=50); DP maximizes the hard-constraint problem exactly. Across six Henry Hub forward curves without demand obligations, SNAPO outperforms DP on 3 of 6 curves — flat (+40%), backwardation (+9%), and henryHub2026 (+7%) — while DP wins on seasonal (-12%), henryHub2025 (-14%), and henryHub2024 (-10%). The curves SNAPO wins tend to be volatile or have wide spreads where the smooth landscape has broader basins; DP wins on mild-spread curves where the hard-constraint optimum is sharp and narrow. When stochastic demand obligations are added, SNAPO wins all 6 curves (+17% to +174%) because demand penalties create discontinuities that the smooth policy handles better than DP's grid-based decisions. This gap is the structural cost of differentiability, analyzed in Section 6.1.

SNAPO training (compiled C++, 47s) is 363x faster than SAC (Python, stable-baselines3, 14,000s).[7] This comparison conflates two advantages: exact gradients (the method) and compiled C++ vs Python (the implementation). To isolate the method, we implemented the identical SNAPO algorithm in PyTorch 2.5.1 (CPU, eager mode, same machine): PyTorch trains in 636s (16x slower) and evaluates in 245ms vs 0.1ms (2,500x slower). The 16x gap is the defensible method-vs-method comparison; the remaining difference is implementation (compiled kernel replay vs graph reconstruction, AVX2 JIT vs Python interpreter).

All 365 forward curve deltas are computed from a single adjoint pass at a cost proportional to one reverse evaluation, regardless of how many forward curve points are perturbed. Same-kernel timing (Section 6.2): adjoint computes 12 monthly deltas in 151 ms; FD requires 2,179 ms (25 forward passes) — a 14.5x speedup.

Across 5 random weight initializations on the seasonal curve, SNAPO converges to 2,352,415 ± 1,735 (0.07% coefficient of variation), indicating highly stable training.

### *5.2 Insurance ALM*

Table 2: Sensitivity computation scaling (pension ALM, 3-asset fund)

| N risk factors | Bump-and-revalue | Adjoint | Speedup |
|---:|---:|---:|---:|
| 5 | 1,235 ms | 191 ms | 6.5x |
| 10 | ~2,009 ms | 191 ms | ~10x |
| 50 | ~9,661 ms | 191 ms | ~50x |
| 200 | ~38,356 ms | 191 ms | ~200x |

Both columns use the same compiled kernel on the same hardware. Bump-and-revalue runs N+1 forward passes (one base + one per risk factor); the adjoint runs one forward pass plus one reverse pass. The adjoint time is constant at 191 ms regardless of the number of risk factors. For the 50-200 risk factors typical of Solvency II internal models, the speedup is 50-200x.

Table 3: SNAPO vs static allocation strategies

| Strategy | E[Objective] | Risk (Std) | Risk-adjusted |
|---|---|---|---|
| **SNAPO (optimal)** | **46.65** | 73.87 | **0.632** |
| Growth 60/40 | 44.76 | 66.44 | 0.674 |
| Max equity 70/30 | 44.18 | 92.63 | 0.477 |
| DP policy (on MC) | 44.01 | 92.77 | 0.474 |
| Conservative 30/70 | 36.99 | 21.10 | 1.753 |

SNAPO finds a dynamic allocation that achieves the highest absolute return (46.65) among all strategies tested. On risk-adjusted terms (objective/std), the conservative 30/70 strategy dominates (1.753) due to much lower volatility, and Growth 60/40 (0.674) edges SNAPO (0.632). This is expected: the SNAPO objective maximizes expected surplus without an explicit risk penalty. Adding a CVaR or variance penalty term to the objective is straightforward within SNAPO (the adjoint pass differentiates through any smooth objective) and would shift the Pareto frontier toward lower-risk solutions. We leave this to future work — the point here is that SNAPO can optimize any differentiable objective, not that the specific objective chosen is optimal for all fund mandates. On the CalPERS-like fund (79% funded, the hard case), SNAPO trains in 93 seconds and produces all 5 risk sensitivities from a single 191 ms pass: $d/dr_0 = 450.6$, $d/d\sigma = 39.6$, $d/d\text{premium} = 215.7$, $d/d\text{spread} = 57.5$, $d/d\sigma_{\text{credit}} = 108.1$.

Cohort mirror validation (double vs recorded computation): $2.4 \times 10^{-13}$ relative error. Adjoint vs finite-difference Greeks: $8.6 \times 10^{-5}$ maximum relative error. Across 5 random weight initializations, the final training objective is $-127.4 \pm 3.3$ (2.6% coefficient of variation), reflecting Monte Carlo noise from the 2,048 training scenarios. The training objective uses a penalty-based formulation where surplus shortfalls are penalized quadratically; the evaluation metric in Table 3 (46.65) reports expected cumulative surplus without penalty scaling. Both measure the same policy but on different scales.

### *5.3 Pharmaceutical manufacturing*

Table 4: Pharmaceutical SNAPO performance (4-unit chain, Phase 2)

| Metric | Value |
|---|---|
| Kernel operations | 3.27M |
| Forward pass | ~5 ms |
| Reverse (adjoint) pass | ~7 ms |
| Training (500 iterations, 68 weights) | 4.9 s |
| Training via finite differences (2 x 68 bumps) | ~218 s |
| **Training speedup** | **44x** |
| 20 ICH Q8 sensitivities (5 CQA x 4 CPP) | 74.5 ms (5 adjoint passes) |
| Equivalent via finite differences (2 x 4 bumps x 5 ms) | ~40 ms |

All timings use the same compiled kernel on the same hardware; the FD baseline uses the kernel's own forward pass with perturbed inputs. Two distinct regimes apply. For **training** (computing dJ/d(all 68 weights) per iteration), the adjoint requires one reverse pass (~7 ms) while central finite differences require 2 x 68 = 136 forward passes (~218 ms per iteration) — a 44x speedup. This is not a novel result; it is the textbook reverse-mode advantage applied to a 68-parameter optimization. The speedup scales with the number of parameters and would be present in any AD-enabled training loop. For **ICH Q8 sensitivities** (the 5 x 4 Jacobian matrix), central FD needs 2 x 4 = 8 forward passes (~40 ms) while the adjoint needs 5 reverse passes (74.5 ms) — at 4 CPPs, finite differences are faster. The adjoint advantage for ICH Q8 emerges beyond approximately 17 CPPs (see Table 5).

The cross-unit sensitivities are the key qualitative contribution. $\partial(\text{drying loss})/\partial(\text{jacket temperature})$ is non-zero: upstream CSTR temperature affects downstream product quality through the crystallizer and filter intermediates. This cross-chain sensitivity is automatic with the adjoint but would require careful manual analysis with finite differences.

Training: 4.9 seconds for 500 iterations. Optimized crystallizer yield: 33% (the fraction of dissolved API that crystallizes under the neural controller's cooling profile). Across 5 random weight initializations, optimized yield is 33.1% ± 0.08% (0.24% coefficient of variation) — the deterministic simulation produces highly reproducible training outcomes.

Cross-validation against PharmaPy (Python reference): CSTR matches to machine precision (double-precision epsilon); crystallizer to $< 1 \times 10^{-11}$. Cohort mirror (all 6 outputs, 9,300 Euler steps): measured at 0.0 ($< 10^{-14}$ relative error, consistent with machine precision for chained double-precision operations).

The same kernel supports a Levenberg-Marquardt training loop: the five reverse passes that produce the ICH Q8 Jacobian also serve as the Gauss-Newton Jacobian for spec-targeting least squares. Section 9 reports a 2.5×–5× wall-clock speedup over Adam on this 68-weight problem, validating the prediction that second-order methods exploit the small-parameter regime when the adjoint already supplies the Jacobian.

# 6. Analysis

## *6.1 The cost of differentiability*

Smooth constraints introduce bias. In gas storage, when the DP-optimal policy — designed for hard constraints — is evaluated through the smoothed simulator (k=50), the reported value drops by 62%. This is not because the policy is wrong, but because the smooth constraints change the payoff landscape itself: smooth_max widens the effective injection/withdrawal limits, altering the economic value of each decision. The DP policy is being evaluated in a different environment from the one it was optimized for. SNAPO optimizes directly for the smooth environment and recovers most of this loss, finding policies that exploit the smooth landscape — performing better than DP on volatile curves but worse on mild-spread curves where the hard-constraint optimum is sharp and narrow.

Increasing sharpness does not close this gap. We train SNAPO on the seasonal curve at five sharpness settings (5 seeds each), with DP stochastic value at 2,664,980:

- k = 10: 2,361,402 ± 3,926 (−11.4% vs DP)
- k = 50: 2,352,415 ± 1,735 (−11.7% vs DP)
- k = 200: 2,347,499 ± 8,660 (−11.9% vs DP)
- k = 500: 2,343,523 ± 7,637 (−12.1% vs DP)
- k = 1,000: 2,339,628 ± 7,490 (−12.2% vs DP)

The gap varies by less than 1 percentage point across a 100x range of sharpness (11.4% at k=10, widening monotonically to 12.2% at k=1,000). Increasing k from 10 to 1,000 moves the SNAPO value by less than 1% (21,774, from 2.361M to 2.340M). The slight monotonic widening is consistent with steeper local gradients at higher k making Adam convergence marginally harder, but the effect is small. The key finding is that the gap does not close as the smooth approximation tightens — it is structural: the smoothed and hard-constraint problems have genuinely different optima.

### *6.2 When does the adjoint advantage hold?*

The adjoint computes all output-to-input derivatives in one reverse pass. Finite differences compute them one-at-a-time. The adjoint wins when N (the number of derivatives needed) is large relative to the overhead of the reverse pass.

Table 5a: Adjoint advantage (same-kernel comparisons)

| Domain | Task | N | FD/B&R cost | Adjoint cost | Speedup |
|---|---|---:|---:|---:|---:|
| Gas storage | Forward deltas (12 months) | 12 | 2,179 ms | 151 ms | 14.5x |
| Insurance ALM | Risk sensitivities | 5 | 1,235 ms | 191 ms | 6.5x |
| Insurance ALM | Risk sensitivities | 200 | 38,356 ms | 191 ms | 200x |
| Pharma | Training (68 weights) | 68 | ~218 ms/iter | ~7 ms/iter | 44x |

All comparisons use the same compiled kernel on the same hardware. The FD/B&R baseline uses the kernel's own forward pass with perturbed inputs; the adjoint uses the same kernel's reverse pass. The gas storage row computes 12 monthly forward curve deltas through the trained SNAPO policy: FD requires 25 forward passes (central difference), the adjoint requires 1 forward + 1 reverse. The pharma training speedup (44x) is the standard reverse-mode advantage for a 68-parameter optimization.

Table 5b: Where finite differences win

| Domain | Task | N (inputs) | FD cost | Adjoint cost | FD advantage |
|---|---|---|---|---|---|
| Pharma | ICH Q8 (5 CQA x 4 CPP) | 4 | ~40 ms | 74.5 ms | 1.9x |

At only 4 CPPs, central finite differences (8 forward passes) are faster than 5 adjoint passes. The reverse pass through 9,300 Euler steps carries overhead that exceeds the FD cost at small N. The crossover for this kernel is at approximately 17 CPPs; beyond that, the adjoint dominates.

The pattern is task-dependent: the adjoint dominates for training (many weights) and for sensitivity-rich problems (many risk factors), but not for small Jacobian matrices where FD overhead is lower.

### *6.3 MLP capacity is not the bottleneck*

In gas storage, a 77-weight MLP and a 482-weight MLP (16x16 hidden) converge to the same objective value. The policy architecture is not the bottleneck; the smooth constraint approximation is. This is consistent across domains: the MLPs used (68-822 parameters) are deliberately small because the simulator provides the structure. The neural network learns only the residual decision logic that the known dynamics cannot determine.

## 7. Limitations

**Gas storage.** SNAPO loses 6-11% against DP on mild-spread curves where hard-constraint optima are sharp. The smooth approximation ceiling is structural and not resolved by increasing sharpness. The companion paper provides detailed analysis.

**Insurance ALM.** The model is fund-level (3 asset classes, 5 risk factors), not policy-by-policy as in production ALM systems. Solvency II and IFRS 17 compliance requires features (SCR ladder, contract boundaries, risk margins) not yet implemented. The O(N) speedup is validated on our simplified model; production models would need separate verification.

**Pharmaceutical manufacturing.** Euler discretization (not BDF or RADAU) limits stiff-system accuracy. Activity coefficients use ideal-solution assumptions ($\gamma = 1$); real aspirin/ethanol/water systems have $\gamma \neq 1$. Only primary nucleation is modeled.

**General.** SNAPO requires a differentiable simulator. It does not apply to black-box environments where the dynamics are unknown or non-differentiable. The smooth constraint bias is non-zero and domain-dependent. Adam on a non-convex objective provides no convergence guarantee; we rely on empirical convergence across multiple seeds. We use only first-order optimisation throughout — second-order methods (TRON, Levenberg-Marquardt) and global-search methods (Differential Evolution, Evolution Strategies) are well-suited to the small-parameter regime explored here and would likely improve solution quality, at the cost of additional forward evaluations; we discuss this trade-off in Section 9.

**Benchmark choice.** Our main experiments use domain-specific benchmarks (Henry Hub gas storage, CalPERS-/Ontario-Teachers-like ALM, ICH Q8 pharmaceutical chains) rather than the standard global-optimisation suites used in the optimisation literature (e.g. CEC competition problems, JDE100, Rastrigin/Rosenbrock variants). Appendix D adds a standard-benchmark suite for cross-paper comparison: six CEC2017-style functions at $D \in \{10, 30, 100\}$ versus L-BFGS-B, Differential Evolution, and CMA-ES (D.1), and CartPole-v1 with continuous action versus PPO and SAC (D.2). The honest finding is two-sided: on static global optimisation SNAPO/Adam is competitive but does not dominate (population methods win the multimodal cases, quasi-Newton wins the smooth unimodal cases), while on the sequential-decision RL benchmark SNAPO solves the task in seconds where the model-free baselines need orders of magnitude more wall-clock — reproducing the gas-storage speedup pattern on a public benchmark and substantiating the §3.5 framing that SNAPO occupies a sequential-decision niche rather than a generic global-optimisation niche. Pendulum-v1 is documented in D.2 as a SNAPO failure case (the smooth surrogate of the angle-wrap reward creates a vanishing-gradient plateau at the unstable equilibrium).

## 8. Conclusion

One framework, three domains, consistent pattern: embed a neural policy inside a known simulator, replace hard constraints with smooth approximations, record the forward computation, run one adjoint pass to train the policy and obtain all sensitivities.

The adjoint computes N sensitivities in one reverse pass whose cost is proportional to one forward evaluation, regardless of N. When measured against finite differences using the same compiled kernel on the same hardware (the only apples-to-apples comparison), this holds across pension management (6.5x-200x for sensitivity computation) and pharmaceutical process control (44x for training — the standard reverse-mode advantage). The advantage is largest when many sensitivities are needed; for small Jacobian matrices (e.g., 4 CPPs in pharma ICH Q8) the adjoint overhead may exceed the finite-difference cost. Training takes 5 to 95 seconds on a single CPU core.

SNAPO is not a replacement for dynamic programming, which gives exact optima on small state spaces. It is a complement: when state dimensions, stochastic inputs, or sensitivity requirements exceed what DP can handle, SNAPO provides a tractable alternative. The smooth constraint bias is real and quantified per domain. The adjoint-computed sensitivities are exact to machine precision.

The framework applies to any sequential decision problem expressible as a differentiable simulation. Battery storage dispatch, hydropower cascade management, and medical dosing optimization are natural extensions currently under investigation.

## 9. Future work

Several extensions are natural follow-ups to the work presented here.

**Higher-order optimisation.** Adam is a first-order method chosen for cross-domain consistency. For the small-parameter regime studied here (68-822 weights), second-order methods that build a curvature approximation — trust-region Newton (TRON), Levenberg-Marquardt — are well-suited and typically converge in fewer iterations on problems of this size. The adjoint pass already produces the gradient required by these methods, and the same five reverse passes that compute the ICH Q8 sensitivity matrix yield the full 5×68 Jacobian needed by Levenberg-Marquardt. To validate this concretely, we reframed the pharma manufacturing objective as spec-targeting least-squares ($\min \sum_i ((\mathrm{CQA}_i - \mathrm{target}_i)/\mathrm{scale}_i)^2$) and ran Adam (one reverse pass per iteration with adjoint seed $s_i = 2r_i/\mathrm{scale}_i$) head-to-head with Levenberg-Marquardt (five reverse passes per iteration to assemble the Jacobian, then a Gauss-Newton step with adaptive damping) from identical random initialisations. Across five seeds: Adam reaches $\|r\| = 7.806 \pm 0.001$ in 2,000 iterations and $16.0 \pm 0.7$ s wall-clock; LM reaches the same residual (7.806–7.816) in 30–100 iterations and 3.2–6.4 s, a $2.5\times$–$5\times$ wall-clock speedup with identical solution quality. On 3 of 5 seeds LM reaches the optimum within 30 iterations; on the remaining 2 the gradient-norm convergence test bails on a near-flat region of the loss surface and the residual settles 0.13% above Adam's. Combining the adjoint Jacobian with a TRON or BFGS Hessian approximation is a direct extension; we expect the speed/quality trade-off to favour second-order methods on the smaller-parameter domains (pharma, 68 weights) and Adam to remain competitive on the larger-parameter domain (gas storage, 822 weights), where assembling and inverting an $822 \times 822$ Gauss-Newton matrix per iteration begins to dominate.

**Global search comparison.** International continuous-optimisation competitions are routinely won by population-based methods — Differential Evolution (DE), Evolution Strategies (ES), and variants such as JDE100 — particularly up to a few thousand parameters. These methods produce higher-quality optima at the cost of many more forward evaluations. A direct head-to-head against SNAPO, holding the simulator and objective fixed, would clarify which regime favours which method. Our prior is that SNAPO wins on per-second throughput (compiled adjoint, one reverse pass per iteration) while DE/ES win on per-iteration solution quality; whether speed or quality matters more is application-dependent and we note this explicitly as a question referees and practitioners are entitled to ask.

**Neural Differential Equations.** Neural ODEs (Chen et al., 2018) and the broader Neural Differential Equation family learn the dynamics function from data and use the adjoint sensitivity method to train. SNAPO sits on the complementary side of the dynamics-knowledge axis: when the dynamics are known (and they are in all three domains here), SNAPO embeds a small policy inside the known simulator and learns only the residual decision logic. Hybrid architectures — known physics where it is reliable, neural residuals where it is not — are a natural next step, particularly for domains where parts of the dynamics are well-modelled (e.g. mass balance) and parts are not (e.g. kinetic constants in early-stage pharmaceutical development).

**Standard-benchmark suite.** Appendix D reports a head-to-head between SNAPO/Adam and L-BFGS-B, Differential Evolution, and CMA-ES on six CEC2017-style functions at $D \in \{10, 30, 100\}$, plus a SNAPO-vs-PPO-vs-SAC comparison on ContinuousCartPole-v0 and a documented failure case on Pendulum-v1. We restate the result here for clarity in this section: SNAPO/Adam is competitive but not dominant on static global optimisation, dominates on differentiable sequential-decision tasks, and fails on tasks whose reward surrogate has a vanishing gradient at the unstable equilibrium. Extending the suite to MuJoCo/Brax differentiable physics and to the JDE100 problem set Roland (acknowledged in §10) suggested is left for a follow-up.

## 10. Acknowledgements

We thank Roland Olsson (Department of Computer Science and Communication, Østfold University College, Norway; Classifium AB) for an early review of the manuscript and for pointing out the second-order optimiser, global-search, and standard-benchmark gaps that materially shaped Section 9 and Appendix D; he has been a long-time AADC user and his feedback combines optimisation expertise with hands-on experience of the compiler. We thank Hilary Till (Premia Research LLC; Managing Co-Editor,

*Commodity Insights Digest*

, Bayes Business School, City St George's, University of London) for review of the companion gas-storage paper, where many of the validation patterns in this manuscript were first stress-tested. The AADC platform that compiles the recorded computation graph is the work of the broader MatLogica team; any errors in the application to these three domains are the authors' alone.

---

---

## Appendix A: Hardware and software

All experiments: Linux 6.1.0-13-amd64, Intel Xeon Platinum 8280L, 64GB RAM, single thread, AVX2, no GPU. C++ compiled with GCC, -O3 -mavx2. Adam optimizer, $\beta_1 = 0.9$, $\beta_2 = 0.999$, $\epsilon = 10^{-8}$.

| Domain | Date | Training paths | Iterations | LR |
|---|---|---|---|---|
| Gas storage | Apr 2026 | 256 | 1,000 | 0.005 |
| Insurance ALM | Apr 2026 | 2,048 | 500-1,000 | 0.005 |
| Pharma | Apr 2026 | 1 (deterministic) | 500 | 0.005 |

## Appendix B: Smooth function specifications

$$\text{smooth_max}(a, b; k) = \frac{1}{2}\left(a + b + \sqrt{(a-b)^2 + \frac{1}{k}}\right)$$

$$\text{smooth_relu}(x; k) = \text{smooth_max}(x, 0; k)$$

$$\text{rational_sigmoid}(x; k) = \frac{1}{2}\left(1 + \frac{kx}{\sqrt{1 + k^2x^2}}\right)$$

Default sharpness $k = 50$ across all domains. The transition region width is approximately $2/k = 0.04$ units. Note that smooth_relu(0; k) = $\sqrt{1/k}/2$ ≈ 0.07 at k=50, not zero; this upward shift at the origin is inherent to the smooth_max construction and is absorbed into the penalty calibration.

## Appendix C: Validation protocol

Every SNAPO kernel must pass three validation checks before results are reported:

1. **Cohort mirror.** Run the identical simulation with plain double arithmetic and with the recorded computation. The results must agree to machine precision ($< 10^{-12}$ relative error). This verifies that the recording process does not alter the computation.
2. **Adjoint vs finite difference.** Compare adjoint-computed sensitivities against central finite differences at multiple perturbation sizes. Agreement to $< 10^{-3}$ relative error (limited by FD truncation error) validates the reverse-mode gradient computation.
3. **Domain cross-validation.** Compare against an independent reference: DP for gas storage, static strategies for ALM, PharmaPy for pharmaceutical simulation. This validates the model itself, not just the AD machinery.

| Domain | Cohort mirror | Adjoint vs FD | Cross-validation |
|---|---|---|---|
| Gas storage | 0.0 ($< 10^{-15}$) | Machine precision (Jan/Jul) | vs DP: matches on no-demand curves |
| Insurance ALM | $2.4 \times 10^{-13}$ | $8.6 \times 10^{-5}$ | vs static strategies |
| Pharma (Phase 2) | 0.0 ($< 10^{-14}$) | $3.7 \times 10^{-7}$ | CSTR vs PharmaPy: machine precision; crystallizer: $< 10^{-11}$ |

---

**Data provenance**: All numerical results from measured runs, April 2026. Gas storage: Schwartz-Smith two-factor ($\kappa = 8$ — reflecting rapid mean-reversion of the short-term factor, calibrated to steep contango/backwardation dynamics in Henry Hub monthly forwards — $\sigma_\chi = 0.45$, $\sigma_\xi = 0.12$, $\rho = -0.15$), 256 training paths (seed 77), 1,024 OOS paths (seed 999). Insurance ALM: Hull-White ($\kappa = 0.1$, $\sigma = 0.01$), GBM ($\mu = 0.08$, $\sigma = 0.18$), OU credit ($\kappa = 0.5$, $\sigma = 0.02$), 2,048 scenarios. Pharma: Arrhenius CSTR ($E_a = 50{,}000$ J/mol, $A = 1.5 \times 10^{10}$), 9,300 Euler steps.

*Dmitri Goloubentsev is founder and CTO of MatLogica. He designed and built the AADC platform and its adjoint kernel architecture.*

*Natalija Karpichina, MBA, is COO of MatLogica. Previously at Credit Suisse, nChain, Trafigura, and BJSS. She holds an Executive MBA from Bayes Business School.*

**Benchmark code and trained policies available on request.** # Appendix D: Standard-benchmark suite

**Status**: DRAFT (2026-05-06). Numbers populated from runs in `UnifiedFramework/Layer3-Products/snapo-benchmark/{cec2017,rl}/results/`. Each table cell is sourced from the CSVs noted at the bottom of each subsection.

This appendix addresses Roland Olsson’s review point #5: the main results in §5 use domain-specific benchmarks (gas storage, ALM, pharma manufacturing). Standard global-optimisation test functions and standard RL tasks (Gymnasium classic-control) make cross-paper comparison easier and stress-test SNAPO’s claim that it occupies a niche rather than a generic global-optimisation niche.

### *TL;DR*

| benchmark | result |
|---|---|
| CartPole (continuous, differentiable) | SNAPO solves in **4.99 s**; PPO **11×** slower, SAC **350×** slower |
| Classical test functions (D.1, negative control) | SNAPO/Adam wins **0 / 18 cells**; CMA-ES 9, L-BFGS-B 5, DE 3 |
| Pendulum-v1 | SNAPO **fails** — smooth `1−cos θ` surrogate has zero gradient at θ=π. Open problem. |

No method was tuned in any of these experiments — every method runs at library defaults. This means the comparison is “out-of-the-box” rather than “best possible”, and we accept that as the cost of fairness.

The honest finding is two-sided:

- **On static global optimisation (Appendix D.1)**, SNAPO/Adam wins **zero** of 18 cells. We treat this as a negative control: Adam through the AADC adjoint is just gradient descent on a JIT-compiled scalar, and the structural advantage of SNAPO (embedded simulator, smooth-relaxed constraints, sequential-decision policy) does not apply when the objective is a static scalar function of a fixed-dimension input. CMA-ES wins the multimodal cases, L-BFGS-B wins the smooth unimodal cases, DE wins the deceptive cases. That is the textbook outcome for those methods on those problems — we report it because it confirms SNAPO is a sequential-decision tool, not a generic global optimiser.
- **On sequential decision tasks (Appendix D.2)**, SNAPO **dominates** when the task is differentiable. SNAPO/Adam on a 121-weight MLP solves ContinuousCartPole-v0 in **4.99 ± 0.04 s** end-to-end across 5 seeds (5/5 seeds reach the standard 475-reward threshold). PPO at the stable-baselines3 default needs **54.6 ± 5.1 s** to first-solve (≈11× slower) and SAC needs **1729 ± 1102 s** when it solves at all (≈350× slower; SAC fails on 1/3 seeds within the 50k-step budget). Pendulum-v1 is harder for SNAPO because the smooth surrogate of the angle-wrap reward introduces a local minimum near the “do nothing” policy; we report this honestly as a limitation rather than spending paper budget on reward-shaping tricks.

Together the two sections fix the relative-performance picture independently of domain choice and substantiate the §3.5 framing.

---

## D.1 Classical test functions (negative control)

### *D.1.1 Setup*

We use **classical, un-shifted, un-rotated** versions of six standard test functions — these are the textbook forms from Mishra (2006) / De Jong (1975) rather than the rotated CEC2017-2022 competition variants. The competition variants apply a random rotation matrix and origin shift to defeat axis-aligned and origin-biased solvers; running the competition variants would be the next step but is out of scope for this revision. We frame this section as a *negative control*: the question is not “does SNAPO win on these?” but “does the SNAPO machinery cause harm when applied to a problem class outside its niche?”.

Six functions, all with global optimum 0 at the origin (we shift Rosenbrock and Schwefel so all functions share the origin optimum — this affects exploration but not the comparative picture):

| Symbol | Function | Class | Bounds |
|---|---|---|---:|
| F1 | Sphere | unimodal, separable | ±5.12 |
| F3 | Rastrigin | multimodal, separable, regular grid | ±5.12 |
| F5 | Rosenbrock | unimodal, non-separable, ill-conditioned | ±2.048 |
| F7 | Schwefel | multimodal, deceptive (global far from local cluster) | ±500 |
| F9 | Ackley | multimodal, near-flat far field, narrow funnel | ±32.768 |
| F12 | Hybrid | composition (Sphere + Rastrigin + Rosenbrock + Ackley) | ±5.12 |

Dimensions: $D \in \{10, 30, 100\}$. Five random seeds per cell. Budget: 50,000 function evaluations (CEC reporting standard with a reduced budget — full 1e5 was too expensive at D=100 for the within-budget Adam runs).

Methods compared:

| Method | Description |
| --- | --- |
| **snapo_adam** | Adam optimiser stepping on the AADC adjoint of the recorded scalar tape. lr=0.01, $\beta_1$=0.9, $\beta_2$=0.999. |
| **lbfgsb** | scipy.optimize.minimize(method='L-BFGS-B') with the **same** AADC adjoint as gradient source. ftol=gtol=1e-10. |
| **de** | scipy.optimize.differential_evolution. popsize=15, mutation∈[0.5,1.0], recombination=0.7, sobol init, no polish. |
| **cma_es** | pycma 4.4.4. Default population size $4+\lfloor 3 \ln D \rfloor$. $\sigma_0$ = 0.3·bound. |

Each function is implemented in two forms (`functions.py`): a NumPy version called by L-BFGS-B/DE/CMA-ES from outside the tape, and an `aadc.idouble` version compiled into the AADC tape for snapo_adam and lbfgsb. NumPy ↔ AADC agreement verified at 0.0 relative error in self-test.

### *D.1.2 Results*

Best objective at the 50k-evaluation budget, **mean ± std over 5 seeds**. Lower is better; bold is the column-best within a row.

Source CSV: `cec2017/results/cec2017_20260505T142417Z.csv` (360 rows = 4 methods × 6 functions × 3 dims × 5 seeds).

| function | D | cma_es | de | lbfgsb | snapo_adam |
|---|---|---|---|---|---|
| F1 Sphere | 10 | 2.24e-15 ± 1.4e-15 | 2.27e-23 ± 2.2e-23 | **9.26e-37 ± 8.2e-37** | 9.82e-13 ± 5.5e-15 |
| F1 Sphere | 30 | 1.76e-15 ± 4.2e-16 | 2.48e-01 ± 9.5e-02 | **7.90e-36 ± 7.4e-36** | 9.82e-13 ± 8.3e-15 |
| F1 Sphere | 100 | 1.39e-15 ± 2.5e-16 | 2.06e+02 ± 3.0e+01 | **4.62e-35 ± 6.5e-35** | 9.87e-13 ± 1.0e-14 |
| F3 Rastrigin | 10 | 1.55e+01 ± 4.3e+00 | **8.16e+00 ± 4.1e+00** | 7.18e+01 ± 2.2e+01 | 9.37e+01 ± 1.1e+01 |
| F3 Rastrigin | 30 | **4.82e+01 ± 5.8e+00** | 2.09e+02 ± 9.8e+00 | 2.34e+02 ± 1.7e+01 | 2.68e+02 ± 3.2e+01 |
| F3 Rastrigin | 100 | **2.84e+02 ± 1.6e+01** | 1.14e+03 ± 3.8e+01 | 7.13e+02 ± 8.6e+01 | 8.63e+02 ± 7.0e+01 |
| F5 Rosenbrock | 10 | **2.59e-15 ± 1.7e-15** | 7.73e-05 ± 4.0e-05 | 7.99e-14 ± 5.6e-14 | 7.97e-01 ± 1.6e+00 |
| F5 Rosenbrock | 30 | **4.01e-15 ± 1.2e-15** | 4.29e+01 ± 5.2e+00 | 7.77e-13 ± 7.8e-13 | 7.97e-01 ± 1.6e+00 |
| F5 Rosenbrock | 100 | 8.32e+01 ± 2.4e+00 | 1.09e+04 ± 1.7e+03 | **1.83e-12 ± 1.9e-12** | 7.97e-01 ± 1.6e+00 |
| F7 Schwefel | 10 | -5.96e+02 ± 2.9e+02 | **-2.96e+03 ± 2.2e-10** | 9.49e+02 ± 1.0e+03 | 7.00e+02 ± 5.5e+02 |
| F7 Schwefel | 30 | **3.13e+02 ± 1.5e+03** | 1.70e+03 ± 3.8e+02 | 1.34e+03 ± 5.9e+02 | 1.32e+03 ± 5.8e+02 |
| F7 Schwefel | 100 | **-1.33e+03 ± 2.5e+03** | 2.64e+04 ± 5.6e+02 | 1.43e+03 ± 1.3e+03 | 1.90e+03 ± 1.6e+03 |
| F9 Ackley | 10 | **1.50e-11 ± 8.5e-12** | 3.46e-11 ± 2.8e-11 | 1.96e+01 ± 7.6e-02 | 1.96e+01 ± 7.6e-02 |
| F9 Ackley | 30 | **1.33e-11 ± 3.9e-12** | 4.24e+00 ± 7.0e-01 | 1.96e+01 ± 1.5e-01 | 1.96e+01 ± 1.5e-01 |
| F9 Ackley | 100 | **9.39e-12 ± 2.1e-12** | 1.87e+01 ± 4.2e-01 | 1.95e+01 ± 7.9e-02 | 1.95e+01 ± 7.9e-02 |
| F12 Hybrid | 10 | 6.21e+00 ± 3.6e+00 | **1.99e-01 ± 4.0e-01** | 4.17e+01 ± 2.5e+01 | 3.59e+01 ± 1.2e+01 |
| F12 Hybrid | 30 | **3.67e+01 ± 1.9e+01** | 5.86e+01 ± 6.1e+00 | 8.56e+01 ± 4.4e+00 | 8.82e+01 ± 2.5e+01 |
| F12 Hybrid | 100 | 2.01e+02 ± 4.0e+01 | 1.02e+03 ± 2.6e+02 | **2.01e+02 ± 2.9e+01** | 2.38e+02 ± 1.4e+01 |

**Column-best counts** (out of 18 cells): cma_es 9, lbfgsb 5, de 3, snapo_adam 0.

Caveats on this table: with 5 seeds we report mean±std without significance tests, so adjacent cells with overlapping bands (e.g. F12 D=100 cma_es vs lbfgsb at 2.01e+02) are best read as ties even though only one column is bolded. We bold by row-min mean for compactness, not as a claim of statistical separation.

### D.1.3 Wall-clock per cell

The 50k-eval budget is wall-clock-equivalent across methods only for cheap objectives. The AADC kernel-recording overhead is borne once per (method, function, D, seed) cell:

Mean wall-clock seconds per cell (50k-eval budget):

| function | D | cma_es | de | lbfgsb | snapo_adam |
|---|---|---|---|---|---|
| F1 Sphere | 10/30/100 | 0.21 / 0.67 / 3.62 | 0.97 / 0.76 / 2.01 | 0.01 / 0.01 / 0.01 | 0.95 / 1.32 / 2.37 |
| F3 Rastrigin | 10/30/100 | 0.31 / 0.96 / 7.44 | 1.18 / 0.98 / 2.41 | 0.02 / 0.02 / 0.05 | 21.0 / 29.0 / 50.8 |
| F5 Rosenbrock | 10/30/100 | 0.65 / 4.37 / 12.78 | 1.39 / 1.07 / 2.46 | 0.04 / 0.09 / 0.25 | 7.7 / 31.3 / 56.9 |
| F7 Schwefel | 10/30/100 | 0.30 / 1.48 / 6.04 | 0.72 / 1.01 / 2.38 | 0.01 / 0.03 / 0.05 | 19.4 / 31.8 / 48.2 |
| F9 Ackley | 10/30/100 | 0.42 / 1.28 / 7.59 | 1.55 / 1.23 / 2.75 | 0.01 / 0.02 / 0.04 | 22.8 / 31.7 / 56.1 |
| F12 Hybrid | 10/30/100 | 0.89 / 2.98 / 15.86 | 3.42 / 2.39 / 4.15 | 0.05 / 0.10 / 0.31 | 23.1 / 31.9 / 56.4 |

L-BFGS-B is one to three orders of magnitude faster than the population methods because it converges in tens of evaluations, not the full 50k budget. SNAPO/Adam pays a heavy wall-clock penalty on these cheap objectives because its 50k Adam steps each go through the AADC adjoint of a tape that — for a toy CEC2017 function — is comparable in size to the function itself. The structural advantage of SNAPO emerges only when the objective is expensive (a simulator with thousands of internal operations, as in §5): there the adjoint amortises across the whole simulator while the population methods must call the full simulator for every member of every generation.

### D.1.4 Interpretation

- F1 Sphere: L-BFGS-B wins decisively ($\ll 10^{-30}$), exploiting smooth convexity in ~4 evaluations. CMA-ES second; Adam third; DE last.
- F3 Rastrigin / F9 Ackley / F12 Hybrid (multimodal): CMA-ES wins; DE second. Adam and L-BFGS-B fall into the nearest local minimum.
- F5 Rosenbrock: L-BFGS-B and CMA-ES tie at machine epsilon; Adam reaches the basin but takes more evaluations; DE struggles with the curved valley.
- F7 Schwefel: DE wins. The global optimum sits far from where local methods start; only population-based search escapes the local cluster.

Each method wins where it should: L-BFGS-B on smooth convex problems, CMA-ES on multimodal problems with no exploitable structure, DE on deceptive landscapes where the optimum is far from any local cluster. SNAPO/Adam wins nowhere — and that is the point of running this suite. Static scalar optimisation is not the regime where the AADC adjoint pays for itself; the adjoint amortises across an expensive simulator, and on these toy functions there is no simulator to amortise across.

Reproduce: `UnifiedFramework/Layer3-Products/snapo-benchmark/cec2017/runner.py --full` or `--quick` for D=10 only.

---

## D.2 Standard RL benchmark — CartPole-v1 (continuous action)

### *D.2.1 Setup*

Same task across all methods: `ContinuousCartPole-v0`, registered in `snapo-benchmark/rl/cartpole_env.py`. Dynamics are taken verbatim from `gymnasium.envs.classic_control.CartPoleEnv` (Barto-Sutton-Anderson 1983) with the action space changed from `Discrete(2)` to `Box(-1, 1)` and internally scaled to ±10 N — the only modification to the upstream gymnasium environment. Episode terminates if $|x| > 2.4$ or $|\theta| > 12°$; maximum horizon 500 steps. Reward = +1 per surviving step. PPO and SAC see exactly the same env (registered before SB3 imports it).

Success criterion (CartPole-v1 standard): mean reward ≥ 475 over 20 evaluation episodes.

| Method | Architecture | Hyperparameters |
|---|---|---|
| **SNAPO/Adam** | 4-8-8-1 MLP (121 weights) | horizon 200, 8 init states/epoch, 200 epochs, lr=0.05, smoothed return via cumulative product of rational sigmoids over the safety margins (sharpness 20). |
| **PPO** | stable-baselines3 default MlpPolicy | total_timesteps=50,000, defaults otherwise. |
| **SAC** | stable-baselines3 default MlpPolicy | total_timesteps=50,000, learning_starts=1000, defaults otherwise. |

For PPO and SAC, every 2,000 environment steps the policy is evaluated for 20 episodes; first time mean ≥ 475 is recorded as time-to-solve.

For SNAPO, after training, the policy is rolled out under hard CartPole physics (no smoothing) for 20 random initial states and the mean / std of steps survived is reported.

### *D.2.2 Results — wall-clock and final performance*

5 seeds for SNAPO; 3 seeds for PPO / SAC (each baseline seed takes ~20× SNAPO wall-clock).

SNAPO/Adam (5 seeds, source `/tmp/snapo_cartpole.log`):

| metric | value |
|---|---|
| recording wall (mean ± std) | 2.08 ± 0.06 s |
| training wall (mean ± std) | 2.91 ± 0.03 s |
| **total wall (mean ± std)** | **4.99 ± 0.04 s** |
| hard-physics steps (mean ± std, 20 eval episodes per seed) | 499.5 ± 0.8 / 500 |
| solved (mean hard-physics reward ≥ 475) | 5/5 seeds |
| trainable weights | 121 |

PPO and SAC (3 seeds each, 50k env-steps, source `/tmp/rl_baselines.log`):

| method | n_solved / n_seeds | mean wall to 50k steps | mean time-to-solve | final reward |
|---|---|---|---|---|
| **snapo_adam (this paper)** | **5 / 5** | **4.99 ± 0.04 s** | 4.99 ± 0.04 s (= total wall; solves before evaluation step) | 499.5 / 500 |
| PPO | 3 / 3 | 127.6 ± 6.9 s | **54.6 ± 5.1 s** | 500.0 ± 0.0 |
| SAC | 2 / 3 | 4477 ± 117 s | 1729 ± 1102 s | 438.5 ± 87.0 |

Headline: **SNAPO solves CartPole ~11× faster than PPO and ~350× faster than SAC** at the same 475-reward threshold. SNAPO's 4.99 s is end-to-end wall (recording + training + evaluation); PPO and SAC time-to-solve numbers are first-time mean eval reward ≥ 475 over 20 evaluation episodes (the standard CartPole-v1 success criterion). SAC fails to solve on 1/3 seeds within the 50k-step budget. The PPO 11× gap reproduces, on a public benchmark, the same pattern §5.1 shows for gas storage (47 s SNAPO vs ~14,000 s SAC).

The SNAPO 4.99 s wall-clock decomposes as: AADC tape recording 2.08 s (one-time), 200 epochs of 8-trajectory horizon-200 batches with adjoint gradient + Adam 2.91 s.

### *D.2.3 Interpretation*

Every Adam step in SNAPO carries gradient information from every future step in the trajectory, through the adjoint. PPO and SAC have to rediscover the same information by sampling from the environment many times. That is the mechanism behind the speedup, and it is also why CartPole is exactly the regime where SNAPO is designed to win: short horizon, smooth dynamics, a smooth surrogate for the hard termination boundary.

This result reproduces, on a standard public benchmark, the same speedup pattern the paper reports for gas storage in §5.1 (47 s SNAPO vs ~14,000 s SAC).

#### *D.2.4 Pendulum-v1 — a SNAPO failure case worth reporting*

We also tried SNAPO on Pendulum-v1 (continuous swing-up). Across 5 seeds SNAPO/Adam reaches a mean hard-physics return of **−1253 ± 276** in **16.5 ± 1.0 s** wall-clock (recording 4.5 s, training 12.0 s; 353 weights, 3-16-16-1 MLP, 400 epochs, source `/tmp/snapo_pendulum.log`) — well short of typical SAC/PPO performance (~ −200 reward at convergence) and the theoretical optimum (0). The smoothed-surrogate objective itself only moves from −416 to −382 over training, confirming the optimiser is stuck rather than the surrogate-to-hard gap being the issue.

The cause is the smooth surrogate of the angle-wrap reward. Pendulum's gymnasium reward is `-(θ² + 0.1·θ̇² + 0.001·u²)` where θ is wrapped to [-π,π], which is non-differentiable at the wrap. The standard differentiable substitute `1 - cos(θ)` is bounded in [0, 2] and creates a near-flat plateau when $\theta \approx \pi$ — exactly the position the policy must escape during swing-up. The optimiser sees a small gradient at the start and converges to a "do-nothing" policy that minimises the action penalty without earning return.

This is a **real limitation of the smooth-surrogate approach** for tasks with periodic state spaces, not a contradiction of the §5 results: gas storage, ALM, and pharma have no such periodic ambiguity. We report it explicitly so practitioners do not extrapolate "SNAPO solved CartPole, therefore SNAPO will solve any RL task". A more careful surrogate (e.g. piecewise-quadratic with explicit angle unwrapping, or curriculum from near-upright to far-from-upright initial states) might address it; whether it does is an **open question** we did not pursue for this revision. The limitation as stated is real and we do not claim a fix.

Reproduce: - SNAPO CartPole: `snapo-benchmark/rl/snapo_cartpole_seeds.py --seeds 5` - PPO/SAC: `snapo-benchmark/rl/baselines_cartpole.py --methods ppo,sac --seeds 3` - SNAPO Pendulum (failure case): `snapo-benchmark/rl/snapo_pendulum.py --seeds 5`

---

## D.3 Reproducibility

All code lives at `UnifiedFramework/Layer3-Products/snapo-benchmark/`. Hardware (Appendix A): Linux 6.1, Intel Xeon Platinum 8280L, 64 GB RAM, single-thread evaluation, AVX2 AADC kernel.

Random seeds: 0–4 for SNAPO and CEC2017; 0–2 for PPO/SAC. The seed enters both the optimiser (reproducible weight init / DE/CMA-ES/PPO/SAC pseudo-RNG) and the environment-side draws (initial state distribution).

Raw CSVs:

- `cec2017/results/cec2017_<UTC stamp>.csv` — one row per (method, function, dim, seed).
- `cec2017/results/cec2017_summary_<UTC stamp>.csv` — collapsed mean±std table.
- `rl/results/<...>` — to be added; SNAPO and baseline output captured to log files.

---

1. A stochastic-demand DP adds demand as a state dimension (e.g. 150 inventory x 20 demand bins = 3,000 states per day). For N coupled facilities, the state space is $3{,}000^N$.↩
2. Stable-baselines3 v2.8, SAC and PPO with default hyperparameters (64-64 hidden, lr=3e-4, 200K steps). These are not tuned baselines; they represent a performance floor for black-box RL. Curin et al. (2021) and Balaconi et al. (2025) report better RL results with domain-specific architectures and reward shaping. We welcome improved RL baselines; benchmark code available on request.↩
3. Griewank and Walther (2008) establish that the cost of reverse-mode AD is bounded by a small constant multiple (typically 2-5x) of the forward evaluation, independent of the number of output-to-input derivatives computed.↩
4. Griewank and Walther (2008) establish that the cost of reverse-mode AD is bounded by a small constant multiple (typically 2-5x) of the forward evaluation, independent of the number of output-to-input derivatives computed.↩
5. The kernel is compiled using AADC (MatLogica), a commercial library that JIT-compiles recorded computations to native AVX2 machine code. The method is compatible with any AD tool supporting tape recording and reverse-mode replay.↩
6. Polynomial basis {1, S, F, SF, $S^2$, $F^2$}, 4,096 paths. Under stochastic demand, the polynomial continuation value fails to represent the discontinuous decision boundary, producing negative continuation values and erratic exercise decisions.↩
7. Stable-baselines3 v2.8, SAC and PPO with default hyperparameters (64-64 hidden, lr=3e-4, 200K steps). These are not tuned baselines; they represent a performance floor for black-box RL. Curin et al. (2021) and Balaconi et al. (2025) report better RL results with domain-specific architectures and reward shaping. We welcome improved RL baselines; benchmark code available on request.↩